\documentclass[Afour,sageh,times]{sagej}

\usepackage{moreverb,url}
\usepackage{natbib}
\usepackage[colorlinks,bookmarksopen,bookmarksnumbered,citecolor=red,urlcolor=red]{hyperref}
\usepackage{amsmath}
\usepackage{amssymb}
\usepackage{mathtools}
\usepackage{balance}

\newcommand\BibTeX{{\rmfamily B\kern-.05em \textsc{i\kern-.025em b}\kern-.08em
T\kern-.1667em\lower.7ex\hbox{E}\kern-.125emX}}

\begin{document}

\runninghead{Horaud and Dornaika}

\title{Hand-Eye Calibration}

\author{Radu Horaud and Fadi Dornaika}

\affiliation{Inria, Montbonnot Saint-Martin, France}
%\affilnum{2}SAGE Publications Ltd, UK}

%\corrauth{Alistair Smith, Sunrise Setting Ltd
%Brixham Laboratory,
%Freshwater Quarry,
%Brixham, Devon,
%TQ5~8BA, UK.}
%
%\email{alistair.smith@sunrise-setting.co.uk}
%
\begin{abstract}
Whenever a sensor is mounted on a robot hand it is important to know the
relationship between the sensor and the hand. The problem of determining this
relationship is referred to as the hand-eye calibration problem. Hand-eye
calibration is important in at least two types of tasks:
(i)~map sensor centered measurements into the robot workspace frame
and (ii)~allow the robot to precisely move the sensor.
In the past some solutions were proposed in the particular case of the sensor
being a TV camera. With almost no exception, all existing solutions attempt to
solve a homogeneous matrix equation of the form 
$AX=XB$.
This paper has the following main contributions.
First we show that there are two possible formulations of the
hand-eye calibration problem. One formulation is the classical one that we
just mentioned. A second formulation takes the form of the following
homogeneous matrix equation:
$MY=M'YB$.
The advantage of the latter formulation is that the extrinsic and intrinsic
parameters of the camera need not be made explicit. Indeed, this formulation
directly uses the 3$\times$4 perspective matrices ($M$ and $M'$) associated
with two positions of the camera with respect to the calibration frame.
Moreover, this formulation together with the classical one cover a wider 
range of camera-based sensors to be calibrated
with respect to the robot hand: single scan-line cameras, stereo heads,
range finders, etc. 
Second, we develop a common mathematical framework to solve for the hand-eye
calibration problem using either of the two formulations. We represent
rotation by a unit quaternion. We present two methods, (i)~a 
closed-form solution for solving for rotation using unit quaternions 
and then solving for translation and (ii)~a
non-linear technique for simultaneously solving for rotation and translation.
Third, we perform a stability analysis both for our two methods and for the
classical linear method developed by \citep{TsaiLenz89}. This
analysis allows the comparison of the three methods. In the light of this
comparison, the non-linear optimization method, that solves for rotation and
translation simultaneously, seems to be the most robust one with respect
to noise and to measurement errors.

\end{abstract}

%\keywords{Class file, \LaTeXe, \textit{SAGE Publications}}

\maketitle

\section{Introduction}
\label{section:Introduction}
Whenever a sensor is mounted on a robot hand it is important to know the
relationship between the sensor and the hand. The problem of determining this
relationship is referred to as the hand-eye calibration problem. Hand-eye
calibration is important in at least two types of tasks:
\begin{itemize}
\item {\em Map sensor centered measurements into the robot workspace frame}.
Consider for example the task of grasping an object at an unknown location.
First, an object recognition system determines the position and orientation of
the object with respect to the sensor; Second, the object location (position
and orientation) is mapped from the sensor frame to the gripper (hand) frame. The robot may now direct its gripper towards the object and grasp it
\citep{HoraudDornaikaBardLaugier95}.
\item {\em Allow the robot to precisely move the sensor}. This is necessary
for inspecting complex 3-D parts \citep{HoraudMohrLorecki92a}, \citep{HoraudMohrLorecki93},
for reconstructing 3-D scenes with a moving camera \citep{BoufamaMohrVeillon93}, or for
visual servoing (using a sensor inside a control servo loop)
\citep{EspiauChaumetteRives92}.
\end{itemize}

In the past some solutions were proposed in the particular case of the sensor
being a TV camera. 
With almost no exception, all existing solutions attempt to
solve a homogeneous matrix equation of the form (\citep{ShiuAhmad89},
\citep{TsaiLenz89}, \citep{ChouKamel91}, \citep{Chen91}, \citep{Wang92}):
\begin{equation}
A X = X B
\label{eq:AX=XB}
\end{equation}

This paper has the following main contributions.

First we show that there are two possible formulations of the
hand-eye calibration problem. One formulation is the classical one that we
just mentioned. A second formulation takes the form of the following
homogeneous matrix equation:
\begin{equation}
M Y = M' Y B
\label{eq:MY=M'YB}
\end{equation}
The advantage of the latter formulation is that the extrinsic and intrinsic
parameters of the camera need not be made explicit. Indeed, this formulation
directly uses the 3$\times$4 perspective matrices ($M$ and $M'$) associated
with two positions of the camera with respect to the calibration frame.
Moreover, this formulation together with the classical one cover a wider 
range of camera-based sensors to be calibrated
with respect to the robot hand: single scan-line cameras, stereo heads,
range finders, etc. 

Second, we develop a common mathematical framework to solve for the hand-eye
calibration problem using either of the two formulations. We represent
rotation by a unit quaternion. We present two methods, (i)~a 
closed-form solution for solving for rotation using unit quaternions 
and then solving for translation and (ii)~a
non-linear technique for simultaneously solving for rotation and translation.

Third, we perform a stability analysis both for our two methods and for the
classical linear method developed by \citep{TsaiLenz89}. This
analysis allows the comparison of the three methods. In the light of this
comparison, the non-linear optimization method, that solves for rotation and
translation simultaneously,seems to be the most robust one with respect
to noise and to measurement errors.

The remaining of this paper is organized as follows. Section~2 states the
problem of determining the hand-eye geometry from both the standpoints of the
classical formulation and our new formulation. Section~3 overviews the main
approaches that attempted to determine a solution. Section~4 shows that the
newly proposed formulation can be decomposed into two equations.
Section~5 suggests two solutions, one based on the work of 
\citep{FaugerasHebert86} and a new one. Boths these solutions solve 
for the classical and for the new formulations.
Section~6 compares our methods with the well known Tsai-Lenz
method through a stability analysis. Finally, Section~7 describes 
some experimental results and Section~8 provides a short discussion.
Appendix A 
briefly reminds the representation of rotations in terms of unit quaternions, and Appendix B gives a derivation of eq.~(31) using properties outlined in Appendix A.

\section{Problem formulation}

The hand-eye calibration problem consists of computing the rigid
transformation (rotation and translation) between a sensor mounted on a robot
actuator and the actuator itself, i.e., the rigid transformation between the
sensor frame and the actuator frame.

\subsection{The classical formulation} 

\begin{figure}[t!]
\centering
\includegraphics[width=0.5\textwidth]{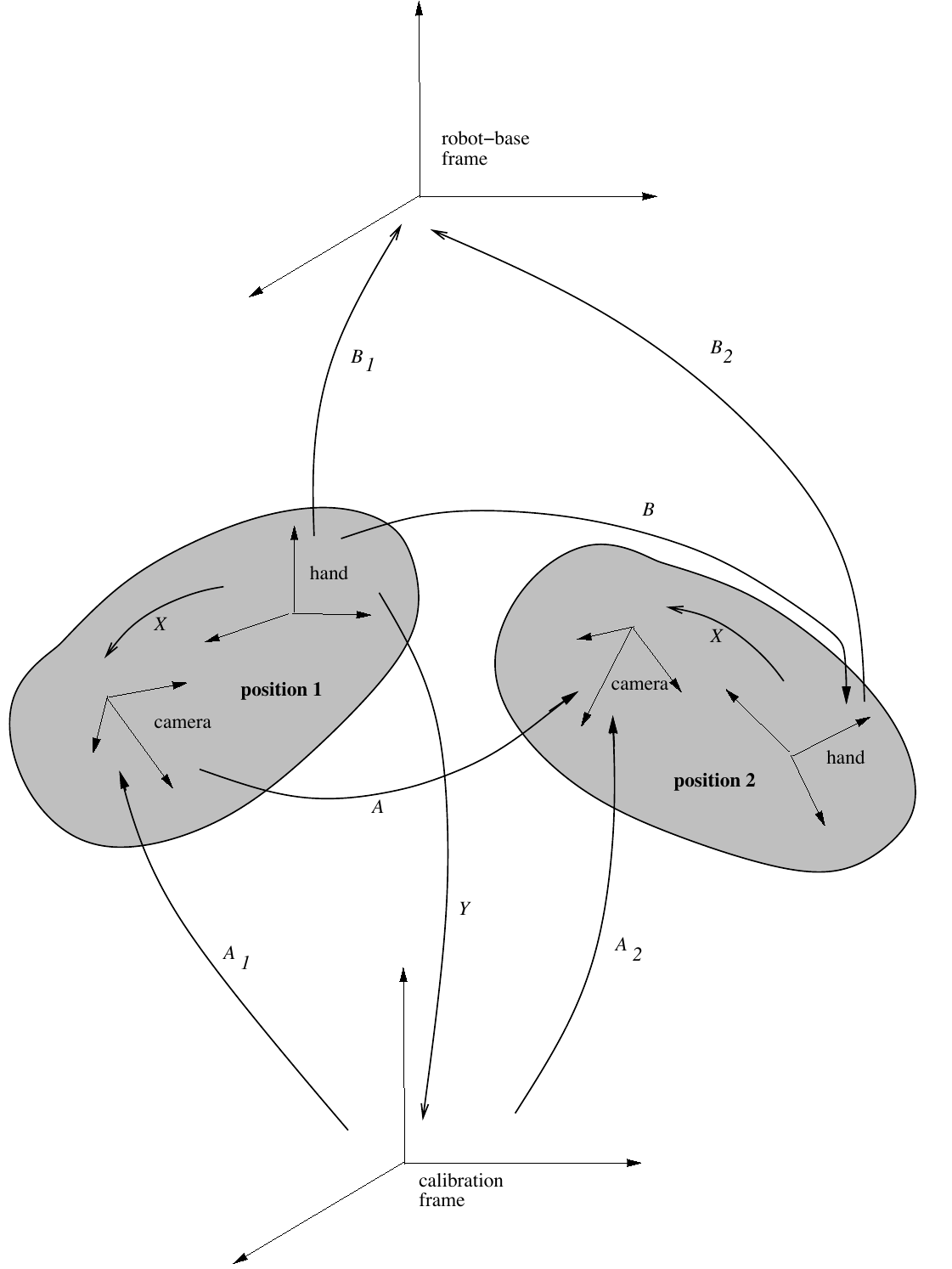}
\caption{\label{fig:general-view} A general view showing two different positions of the hand-eye device.
One is interested to estimate matrix $X$ or, alternatively, matrix $Y$ (see
text).}
\end{figure}

The hand-eye problem is best described on Figure~\ref{fig:general-view}. Let
position 1 and position 2 be two positions of the rigid body formed by a
sensor fixed onto a robot hand and which will be referred to as the
{\em hand-eye device}. Both the sensor and the hand have a Cartesian
frame associated with them. Let $A$ be the transformation between the two
positions of the sensor frame and let $B$ be the transformation between the
two positions of the hand frame. Let $X$ be the transformation between the
hand frame and the sensor frame. $A$, $B$, and $X$ are related by the formula
given by eq.~(\ref{eq:AX=XB}) and they are 4$\times$4 matrices of the form:
\[
A =
\left(
\begin{array}{cc}
   R_{A}    &  t_{A} \\
   0    & 1
\end{array}
\right)
\]
In this expression, $R_{A}$ is a 3$\times$3 orthogonal matrix describing a
rotation, and $ t_{A}$ is a 3-vector describing a translation.

Throughout the paper we adopt the following notation: matrix $T$ ($A$, $B$,
$X$, $Y$, \ldots ) is the
transformation {\em from} frame~$b$ {\em to} frame~$a$:
\[
\left(
\begin{array}{c} x_{a} \\ y_{a} \\ z_{a} \\ 1 \end{array}
\right)
= T
\left(
\begin{array}{c} x_{b} \\ y_{b} \\ z_{b} \\ 1 \end{array}
\right)
\]
where a 3-D point indexed by $a$ is expressed in frame $a$.

In the particular case of a camera-based sensor, the matrix $A$ is obtained by calibrating the camera twice with respect to a
fixed calibrating object and its associated frame, called the
{\em calibration frame}. Let $A_{1}$ and
$A_{2}$ be the transformations from the calibration frame to the camera
frame in its two different positions. We have:
\begin{equation}
A = A_{2} A_{1}^{-1}
\label{eq:A=A2A1}
\end{equation}

The matrix $B$ is obtained by moving the robot hand from position~1 to
position~2. Let $B_{1}$ and $B_{2}$ be the transformations from the hand frame 
in positions 1 and 2, to the robot-base frame. We have:
\begin{equation}
B = B_{2}^{-1} B_{1}
\label{eq:B=B2B1}
\end{equation}

\begin{figure}[t!]
%%\centerline{\psfig{figure=../camera.idraw,height=12cm,width=17cm}}
\centering
\includegraphics[width=0.50\textwidth]{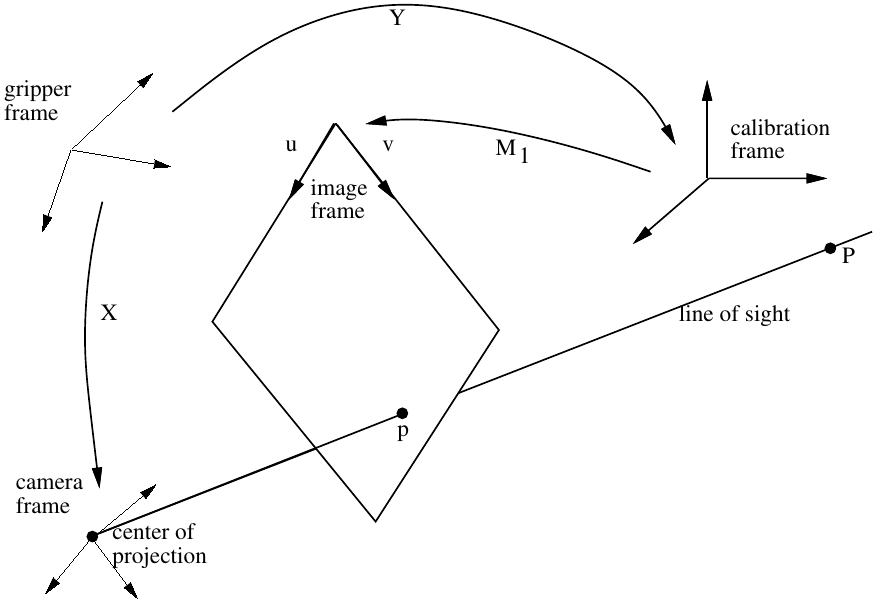}
%\vspace{2cm}
\caption{\label{fig:camera-model} The line of sight passing through the center of projection and the
image point $p$ may well be expressed in the calibration frame, using the
coefficients of the perspective matrix $M_{1}$.}
\end{figure}

\subsection{The new formulation}
\label{section:the-new-formulation}

The previous formulation implies that the camera is calibrated at each
different position $i$ of the hand-eye device. Once the camera is calibrated, its
extrinsic parameters, namely the matrix $A_{i}$ for position~$i$, are made
explicit. This is done by decomposing the 3$\times$4 perspective matrix $M_{i}$,
that is obtained by calibration, into
intrinsic and extrinsic parameters \citep{FaugerasToscani86}, \citep{Tsai87}, \citep{HoraudMonga93}:
\begin{eqnarray}
M_{i} & = & C A_{i} \nonumber \\
& = & \left( \begin{array}{cccc}
        \alpha _{u} & 0 & u_{0} & 0 \\
        0 & \alpha _{v} & v_{0} & 0 \\
        0 & 0 & 1 & 0
       \end{array} \right)
\left(
\begin{array}{cc}
   R_{A}^{i}    &  t_{A}^{i} \\
   0    & 1
\end{array}
\right)
\label{eq:M=CA}
\end{eqnarray}
The parameters $\alpha _{u}$, $\alpha _{v}$, $u_{0}$ and $v_{0}$ describe the
affine transformation between the camera frame and the image frame.
This decomposition assumes that the camera is described by a pin-hole model
and that the optical axis associated with this model is perpendicular to the
image plane. 

The new formulation that we present here avoids the above decomposition.
Let $Y$ be the transformation matrix from the hand frame to the calibration
frame, when the hand-eye device is in position~1. Clearly we have, e.g.,
Figure~\ref{fig:general-view}:
\begin{equation}
X = A_{1} Y
\label{eq:X=A1Y}
\end{equation}
Therefore matrix $Y$ is equivalent to matrix $X$, up to a rigid transformation
$A_1$.  By substituting $X$ given by this last equation and $A$ given by
eq.~(\ref{eq:A=A2A1}) into eq.~(\ref{eq:AX=XB}), we obtain:
\[
A_{2} Y = A_{1} Y B
\]
By pre multiplying the terms of this equality with matrix $C$ and using
eq.~(\ref{eq:M=CA}) with $i=1,2$ we obtain:
\begin{equation}
M_{2} Y = M_{1} Y B
\label{eq:M2Y=M1YB}
\end{equation}
which is equivalent to eq.~(\ref{eq:MY=M'YB}).

In this equation the unknown $Y$ is the transformation from the hand frame to
the calibration frame, e.g., Figure~\ref{fig:general-view}. The latter frame
may well be viewed as the camera frame provided that the 3$\times$4
perspective matrix $M_{1}$ is known. Mathematically, choosing the calibration
frame rather then the camera frame
is equivalent to replacing the 3$\times$4 perspective
matrix $C$ with the more general matrix $M_{1}$. 
The advantage of using $M_{1}$ rather than
$C$ is that one has not to assume a perfect pin hole camera model anymore.
Therefore, problems due to the decomposition of $M_{1}$ into external and 
internal camera parameters, i.e., $M_{i}=CA_{i}$, will disappear.

Referring to Figure~\ref{fig:camera-model}, the projection of a point $P$
onto the image is described by:
\begin{equation}
\left( \begin{array}{l} su \\ sv \\ s \end{array} \right) = 
M_1
\left( \begin{array}{l} x \\ y \\ z \\ 1\end{array} \right)
\end{equation}
or:
\begin{align}
\label{eq:u_def}
u = \frac{m_{11} x + m_{12} y + m_{13} z + m_{14}}{m_{31} x + m_{32} y +
m_{33} z + m_{34}}\\
\label{eq:v_def}
v = \frac{m_{21} x + m_{22} y + m_{23} z + m_{24}}{m_{31} x + m_{32} y +
m_{33} z + m_{34}}
\end{align}
where $x$, $y$, and $z$ are the coordinates of $P$ in the calibration frame,
$u$ and $v$ are the image coordinates of $p$--the projection of $P$, and
the $m_{ij}$'s are the coefficients of $M_{1}$. Notice that these two equations can
be rewritten as:
\begin{align}
\label{eq:u-plan}
(m_{11} - u m_{31}) & \; x + (m_{12} - u m_{32}) \; y \nonumber \\
& +(m_{13} - u m_{33}) \; z = u m_{34} - m_{14} \\
\label{eq:v-plan}
(m_{21} - v m_{31}) & \; x + (m_{22} - v m_{32}) \; y \nonumber \\
& +(m_{23} - v m_{33}) \; z = v m_{34} -  m_{24}
\end{align}

These equations may be interpreted as follows. Given a matrix $M_{1}$ and an
image point $p$, eq.~(\ref{eq:u-plan}) and eq.~(\ref{eq:v-plan}) describe a
line of sight
passing through the center of projection and through $p$. This line is given in
the calibration frame which may well be viewed as the camera frame.

The determination of the hand-eye geometry (matrix $X$ in the classical
formulation or matrix $Y$ in our new formulation) allows one to express any
line of sight associated with an image point $p$ in the hand frame and hence,
in any robot centered frame.
\subsection{Summary}
In practice, the classical and the new formulations summarize as follows. Let $n$
be the number of different positions of the hand-eye device
with respect to a fixed calibration frame. We have:
\begin{enumerate}
\item {\em Classical formulation.} The matrix $X$ is the solution of the following
set of $n-1$ matrix equations:
\begin{equation}
\left\{
\begin{array}{ccc}
	A_{12} X & = & X B_{12} \\
                 & \vdots &	\\
	A_{i-1\; i} X & = & X B_{i-1\; i} \\
		  & \vdots &     \\
	A_{n-1\; n} X & = & X B_{n-1\; n}
\end{array}
\right.
\label{eq:nAXXB}
\end{equation}
where $A_{i-1\; i}$ denotes the transformation between position $i-1$ and position
$i$ of the camera frame and $B_{i-1\; i}$ denotes the transformation between
position $i-1$ and position 
$i$ of the hand frame.
\item {\em New formulation.} The matrix $Y$ is the solution of the following set of
$n-1$ matrix equations:
\begin{equation}
\left\{
\begin{array}{ccc}
	M_{2} Y & = & M_{1} Y B_{12} \\
		& \vdots &     \\
	M_{i} Y & = & M_{1} Y B_{1i} \\
                  & \vdots &     \\
	M_{n} Y & = & M_{1} Y B_{1n}
\end{array}
\right.
\label{eq:nMYNYB}
\end{equation}
where $M_{i}$ is the projective transformation between the calibration 
frame and the camera frame in position $i$ and $B_{1i}$ denotes the
transformation between position 1 and position $i$ of the hand frame.
\end{enumerate}

\section{Previous approaches}
\label{section:previous-approaches}

Previous approaches for solving the hand-eye calibration problem attempted to
solve eq.~(\ref{eq:AX=XB}) ($AX=XB$) by farther decomposing it into two
equations: A matrix equation depending on rotation and a vector equation
depending both on rotation and translation:
\begin{equation}
R_{A} R_{X} = R_{X} R_{B}
\label{eq:AX=XB:rotation}
\end{equation}
and:
\begin{equation}
(R_{A} - I) t_{X} = R_{X} t_{B} - t_{A}
\label{eq:AX=XB:translation}
\end{equation}
In this equation $I$ is the 3$\times$3 identity matrix. 

In order to solve eq.~(\ref{eq:AX=XB:rotation}) one may take advantage of the
particular algebraic and geometric properties of rotation (orthogonal)
matrices. Indeed this equation can be written as:
\begin{equation}
R_{A} =  R_{X} R_{B}  R_{X}^{T}
\end{equation}
which is a similarity transformation since $ R_{X}$ is an orthogonal matrix.
Hence, matrices $R_{A}$ and $R_{B}$ have the same eigenvalues. A well-known
property of a rotation matrix is that is has one of its eigenvalues equal to
1. Let $n_{B}$ be the eigenvector of $R_{B}$ associated with this eigenvalue.
By post multiplying eq.~(\ref{eq:AX=XB:rotation}) with $n_{B}$ we obtain:
\begin{equation*}
R_{A} R_{X} n_{B}  =  R_{X} R_{B} n_{B} =  R_{X} n_{B}
\end{equation*}
and we conclude that $R_{X} n_{B}$ is equal to $n_{A}$, the eigenvector of
$R_{A}$ associated with the unit eigenvalue:
\begin{equation}
n_{A} = R_{X} n_{B}
\label{eq:nA=RXnB}
\end{equation}

To conclude, solving for $AX=XB$ is equivalent to solving for
eq.~(\ref{eq:nA=RXnB}) and for eq.~(\ref{eq:AX=XB:translation}). Solutions
were proposed, among others, by  \citep{ShiuAhmad89}, 
\citep{TsaiLenz89}, \citep{ChouKamel91}, and \citep{Wang92}.
All these authors noticed that at least three positions are necessary in order
to uniquely determine $X$, i.e., $R_{X}$ and $t_{X}$.  \citep{ShiuAhmad89} cast the
rotation determination problem into the problem of solving for 8 linear
equations in 4 unknowns and they used standard linear algebra techniques in
order to obtain a solution.

 \citep{TsaiLenz89}
suggested to represent $R_{X}$ by its unit eigenvector $n_{X}$
and an angle $\theta _{X}$. They noticed that:
\[
n_{X} \cdot (n_{A} -  n_{B}) = 0
\]
and
\[
(n_{A} -  n_{B}) \cdot (n_{A} +  n_{B}) = 0
\]
These expressions allow one to cast eq.~(\ref{eq:nA=RXnB}) into:
\begin{equation}
(n_{A} +  n_{B}) \times n = n_{A} -  n_{B}
\label{eq:linear-rotation}
\end{equation}
with:
\[
n = \left( \tan \frac{\theta _{X}}{2} \right) n_{X}
\]

It is easy to notice that eq.~(\ref{eq:linear-rotation}) is rank deficient and
hence, at least two independent hand-eye motions (at least three positions)
are necessary for determining a unique solution. In the general case of $n$
motions ($n+1$ positions of the hand-eye device relative to the calibration
frame) one may solve for an over constrained set of $2n$ linear equations in 3
unknowns.

 \citep{ChouKamel91}
suggested to represent rotation by a unit quaternion and they
used the singular value decomposition method in order to solve 
for the linear algebra. The idea
of using a unit quaternion is a good one. Unfortunately the authors were not
aware of the closed-form solution that is available in conjunction with
unit quaternions for determining rotation optimally as it was proposed both by \citep{Horn87-quat} and
by \citep{FaugerasHebert86}.

 \citep{Wang92}
suggested three methods that roughly correspond to the solution proposed
by \citep{TsaiLenz89}. Then he compared his best method to the methods proposed by
\citep{ShiuAhmad89} and by \citep{TsaiLenz89}. The conclusions of his comparison are that
the \citep{TsaiLenz89} method yield the best results.

 \citep{Chen91} showed that the hand-eye geometry can be conveniently
described using a screw representation for rotation and translation. This
representation allows a uniqueness analysis. 

All these approaches 
have the following features in common:
\begin{itemize}
\item rotation is decoupled from translation;
\item the solution for rotation is estimated using linear algebra techniques;
\item the solution for translation is estimated using linear algebra as well.
\end{itemize}

Decoupling rotation and translation is certainly a good idea. It leads to
simple numerical solutions. However, in the presence of errors the linear
problem to be solved becomes ill-conditioned and the solution available with
the linear system is not valid.
This is due to the fact that the geometric properties allowing the
linearization of the rotation equation do not hold in the presence of noise.
Errors may be due both to camera calibration inaccuracies and to inexact
knowledge of the robot's kinematic parameters.

\section{Decomposing the new formulation}

In this section we show that the new formulation that we introduced in
section~\ref{section:the-new-formulation} has a mathematical structure that is
identical to the classical formulation. This will allow us to formulate a
unified approach that solves for either of the two formulations.

We start by making explicit the 3$\times$4 perspective matrix $M$ as a
function of intrinsic and extrinsic parameters, i.e., eq.~(\ref{eq:M=CA}):
%\begin{align*}
\begin{multline*}
M = \left( \begin{matrix}
\alpha _u r_{11} + u_{0} r_{31} & \alpha _u r_{12} + u_{0} r_{32} \\
\alpha _v r_{21} + v_{0} r_{31} & \alpha _v r_{22} + v_{0} r_{32} \\
r_{31} & r_{32} 
\end{matrix} \right. \\
\left. 
\begin{matrix}
\alpha _u r_{13} + u_{0} r_{33} & \alpha _u t_{x} +  u_{0} t_{z} \\
\alpha _v r_{23} + v_{0} r_{33} & \alpha _v t_{y} +  v_{0} t_{z} \\
 r_{33} &t_{z}
\end{matrix}
\right)
\end{multline*}
%\end{align*}

Notice that a matrix $M_{i}$ of this form can be written as:
\[
M_{i} = \left( \begin{array}{cc}
N_{i} & n_{i}
\end{array} \right)
\]
where $N_{i}$ is a 3$\times$3 matrix and $n_{i}$ is a 3-vector. One may notice
that in the general case $N_{i}$ has an inverse since the vectors
$(r_{11}\; r_{12}\; r_{13})^{T}$, $(r_{21}\; r_{22}\; r_{23})^{T}$, and
$(r_{31}\; r_{32}\; r_{33})^{T}$ are mutually orthogonal and $\alpha _u \neq
0$, $\alpha _v\neq 0$. With this notation eq.~(\ref{eq:M2Y=M1YB}) may be
decomposed into a matrix equation:
\begin{equation}
N_{2} R_{Y} = N_{1} R_{Y} R_{B}
\label{eq:N2RY=N1RYRB}
\end{equation}
and a vector equation:
\begin{equation}
N_{2} t_{Y} + n_{2} =  N_{1} R_{Y} t_{B} +  N_{1} t_{Y} + n_{1}
\label{eq:N2tY+n2=...}
\end{equation}

Introducing the notation:
\[ N =  N_{1}^{-1} N_{2} \]
eq.~(\ref{eq:N2RY=N1RYRB}) becomes:
\begin{equation}
N  R_{Y} = R_{Y} R_{B} 
\label{eq:NRY=RYRB}
\end{equation}
or:
\[
N = R_{Y} R_{B}  R_{Y}^{T} 
\]
Two properties of $N$ may be easily derived:
\begin{enumerate}
\item $N$ is the product of three rotation matrices, it is therefore a
rotation itself and:
\[	N^{-1} = N^{T}		\]
\item Since $R_{Y}$ is an orthogonal matrix, the above equation defines a
similarity transformation. It follows that $N$ has the same eigenvalues as
$R_{B}$. In particular $R_{B}$ has an eigenvalue equal to 1 and let $n_{B}$ be
the eigenvector associated with this eigenvalue. 
\end{enumerate}

If we denote by $n_{N}$ the
eigenvector of $N$ associated with the unit eigenvalue, then we obtain:
\begin{eqnarray*} N R_{Y} n_{B} & = &  R_{Y} R_{B}  n_{B}\\
                                &  = & R_{Y}n_{B} 
\end{eqnarray*}
and hence we have:
\begin{equation}
n_{N} = R_{Y}n_{B}
\label{eq:nN=RYnB}
\end{equation}
This equation is identical to eq.~(\ref{eq:nA=RXnB}) in the classical
formulation.

By premultiplying eq.~(\ref{eq:N2tY+n2=...}) with $ N_{1}^{-1}$ we obtain:
\begin{equation}
(N - I)  t_{Y} =  R_{Y} t_{B} - t_{N}
\label{eq:N-I...}
\end{equation}
with:
\[ t_{N} =  N_{1}^{-1}(n_{2} - n_{1}) \]
and one may easily notice that this equation is identical to 
eq.~(\ref{eq:AX=XB:translation}) in the classical formulation.

To conclude, the classical formulation decomposes in eq.~(\ref{eq:nA=RXnB})
and in eq.~(\ref{eq:AX=XB:translation}) and, equivalently, the new formulation
decomposes in  eq.~(\ref{eq:nN=RYnB}) and eq.~(\ref{eq:N-I...}).

\section{A unified optimal solution}
\label{section:unified-optimal-solution}
In the previous sections we showed that the classical and the new formulations
are mathematically equivalent. Indeed, the classical formulation, $AX=XB$
decomposes into eqs.~(\ref{eq:nA=RXnB}) and (\ref{eq:AX=XB:translation}):
\begin{eqnarray*}
n_{A} & = & R_{X} n_{B} \\
(R_{A} - I) t_{X} & = & R_{X} t_{B} - t_{A}
\end{eqnarray*}
and the new formulation, $MY=M'YB$ decomposes into eqs.~(\ref{eq:nN=RYnB}) and (\ref{eq:N-I...}):
\begin{eqnarray*}
n_{N} & = & R_{Y}n_{B} \\
(N - I)  t_{Y} & = & R_{Y} t_{B} - t_{N}
\end{eqnarray*}

These two sets of equations are of the form:
\begin{equation}
v' = Rv
\label{eq:general-form-1}
\end{equation}
\begin{equation}
(K-I) t = Rp - p'
\label{eq:general-form-2}
\end{equation}
where $R$ and $t$ are the parameters to be estimated (rotation and
translation), $v'$, $v$, $p'$, $p$ are 3-vectors, $K$ is a 3$\times$3 
orthogonal matrix
and $I$ is the identity matrix.

Eqs.~(\ref{eq:general-form-1}) and (\ref{eq:general-form-2}) are associated
with one motion of the hand-eye device. In order to estimate $R$ and $t$ at
least two such motions are necessary. In the general case of $n$ motions one
may cast the problem of solving 2$n$ such equations into the problem of
minimizing two positive error functions:
\begin{equation}
f_{1}(R) = \sum_{i=1}^{n} \| v'_{i} - R v_{i} \| ^{2}
\label{eq:error-function-rotation-1}
\end{equation}
and
\begin{equation}
f_{2}(R,t) = \sum_{i=1}^{n} \| Rp_{i} - (K_{i}-I)t-p'_{i} \|  ^{2}
\label{eq:error-function-translation-1}
\end{equation}

Therefore, two approaches are possible:
\begin{enumerate}
\item {\em $R$ then $t$.} Rotation is estimated first by 
minimizing $f_{1}$. This
minimization problem has a simple closed-form solution that will be detailed
below. Once the optimal rotation is determined, the minimization of $f_{2}$
over the translational parameters is a linear least-squared problem.
\item {\em $R$ and $t$.} Rotation and translation are estimated simultaneously
by minimizing $f_{1}+f_{2}$. This minimization problem is non-linear but, as
it will be shown below, it provides the most stable solution.
\end{enumerate}
\subsection{Rotation then translation}
In order to minimize $f_{1}$ given by eq.~(\ref{eq:error-function-rotation-1})
we represent rotation by a unit quaternion. With this representation one may
write, (see Appendix A, eq.~(\ref{eq:rotation-quaternion})):
\[	 R v_{i} =  q \ast v_{i} \ast \overline{q} \]
Moreover, using eq.~(\ref{eq:r*q=r2q2}), one may successively write:
\begin{align*}
\| v'_{i} -  & q \ast v_{i} \ast \overline{q} \|^{2}  = 
	\| v'_{i} -  q \ast v_{i} \ast \overline{q} \|^{2}\| q \|^{2} \\ 
 = & \| v'_{i} \ast q - q \ast v_{i} \|^{2} \\
 = & ( Q(v'_{i}) q - W(v_{i}) q )^{T}( Q(v'_{i}) q - W(v_{i}) q ) \\
 = & q^{T}{\cal{A}}_{i}q
\end{align*}
with ${\cal{A}}_{i}$ being a 4$\times$4 positive symmetric matrix:
\[
{\cal{A}}_{i} =  ( Q(v'_{i})  - W(v_{i}) )^{T}( Q(v'_{i}) - W(v_{i}) )
\]
Finally the error function becomes:
\begin{eqnarray}
\label{eq:f1=qTAq}
f_{1}(R) & = & f_{1}(q) \nonumber \\
         & = & \sum_{i=1}^{n} \| v'_{i} -  q \ast v_{i} \ast \overline{q} \|^{2} \nonumber \\
	 & = & \sum_{i=1}^{n} q^{T}{\cal{A}}_{i}q \\
	 & = & q^{T} \left( \sum_{i=1}^{n} {\cal{A}}_{i} \right) q	\nonumber \\
         & = & q^{T} {\cal{A}} q \nonumber
\end{eqnarray}
with ${\cal{A}}=\sum_{i=1}^{n}{\cal{A}}_{i}$ and one has to minimize $f_{1}$ under the
constraint that $q$ must be a unit quaternion. This constrained minimization
problem can be solved using the Lagrange multiplier:
\[ \min_{q} f_{1} =  \min_{q} (q^{T} {\cal{A}} q + \lambda(1-q^{T}q)) \]

By differentiating 
this error function with respect to $q$ one may easily find the
solution in closed form:
\[	{\cal{A}}q = \lambda q 	\]

The unit quaternion minimizing $f_{1}$ is therefore the eigenvector of ${\cal{A}}$
associated with its smallest (positive) eigenvalue. This closed-form solution
was introduced by \citep{FaugerasHebert86} for finding the
best rotation between two sets of 3-D features. 

Once the rotation has been determined, the problem of determining the best
translation becomes a linear least-squares problem that can be easily solved
using standard linear algebra techniques.

\subsection{Rotation and translation}

The problem of estimating rotation and translation simultaneously can
be stated in terms of the following minimization problem:
\[ \min_{q,t}(f_{1} + f_{2}) \]
We have been unable to solve this problem in closed form. One may notice that
the error function to be minimized is a sum of squares of non linear
functions. Because of the special structure of the Jacobian and Hessian
matrices associated with error functions of this type, a number of special
minimization methods have been designed specifically to deal with this case, 
\citep{GillMurrayWright89}. Among these methods, the Levenberg-Marquardt method
and the trust-region method \citep{Fletcher90}, 
\citep{PhongHoraud93a}, \citep{PhongHoraud93b} are good candidates.

Using unit quaternions the error function to be minimized is:
\begin{equation}
\min_{q,t} (f(q,t) + \lambda (1-q^{T}q)^{2})
\label{eq:non-linear-problem}
\end{equation}
with:
\begin{eqnarray*}
f(q,t) & = & \lambda _1 f_{1}(q) + \lambda _2 f_{2}(q,t) \\
       & = & \lambda _1 \sum_{i=1}^{n} \| v'_{i} - q \ast v_{i} \ast \overline{q} \| ^{2} \\
	& + &  \lambda _2 \sum_{i=1}^{n} 
	\| q \ast p_{i} \ast \overline{q} - (K_{i}-I)t-p'_{i} \|  ^{2} 
\end{eqnarray*}
which has the form of sum of squares of non linear functions and $\lambda (1-q^{T}q)^{2}$ is a 
penalty function that guarantees that $q$
(a quaternion) has a module equal to 1. $\lambda_1$ and $\lambda_2$ 
are two weights and $\lambda$ is a real positive number.
High values for $\lambda$ insure that the module of $q$ is closed to 1. In all
our experiments we have: 
\begin{align*}
&\lambda_1  =  \lambda_2 = 1 \\
&\lambda  =  2 \times 10^6
\end{align*}

There are two possibilities for solving the non linear minimization problem of
equation~(\ref{eq:non-linear-problem}). 
The first possibility is to consider it as a classical non linear least squares minimization 
problem and to apply standard non linear optimization techniques, such as Newton's method and 
Newton-like methods \citep{GillMurrayWright89}, \citep{Fletcher90}. 
In the next two sections we give some results obtained with the
Levenberg-Marquardt non linear minimization method as described in \citep{NumericalRecipes}. 

The second possibility is to try to simplify the expression of the error function to be minimized. Using properties associated with quaternions, the error function may indeed be
simplified. We already obtained a simple analytic form for $f_{1}$,
i.e., eq.~(\ref{eq:f1=qTAq}). Similarly, $f_{2}$ simplifies as well.

Indeed, $f_{2}$ is the sum of terms such as:
\[ \| q \ast p_{i} \ast \overline{q} - (K_{i}-I)t-p'_{i} \|  ^{2} \]
and we have:
\begin{align*}
\| q \ast p_{i} \ast \overline{q} & - (K_{i}-I)t-p'_{i} \|^{2} \| q \|^{2} \\
&= \| q \ast p_{i} - (K_{i}-I) t \ast q - p'_{i} \ast q \|^{2}
\end{align*}
Using the matrix representation for quaternion multiplication one can easily
obtain (see Appendix~B for the derivation of this equation):
\begin{align}
\label{eq:f2-quat-transl}
f_{2}(q,t) & = q^{T}(\sum_{i=1}^{n}{\cal{B}}_{i})q + t^{T}(\sum_{i=1}^{n}{\cal{C}}_{i})t
	   + (\sum_{i=1}^{n}\delta _{i})t \nonumber \\
	   &+ (\sum_{i=1}^{n}\varepsilon _{i})Q(q)^{T}W(q)t
\end{align}
The 4$\times$4 matrices ${\cal{B}}_{i}$ and ${\cal{C}}_{i}$, and the 
1$\times$4 vectors $\delta _{i}$ and $\varepsilon _{i}$ are:
\begin{eqnarray*}
{\cal{B}}_{i} & = & (p_{i}^{T}p_{i} + p_{i}'^{T}p_{i}')I - W(p_{i})^{T}Q(p_{i}') -
	    Q(p_{i}')^{T}W(p_{i}) \\
{\cal{C}}_{i} & = & K_{i}^{T}K_{i} - K_{i} - K_{i}^{T} + I \\
\delta _{i} & = & 2p_{i}'^{T}(K_{i} - I) \\
\varepsilon _{i} & = & -2p_{i}^{T}(R_{B_{i}} - I)
\end{eqnarray*}

With the notations: ${\cal{B}}=\sum_{i=1}^{n}{\cal{B}}_{i}$, ${\cal{C}}=\sum_{i=1}^{n}{\cal{C}}_{i}$,
$\delta =\sum_{i=1}^{n}\delta _{i}$, $\varepsilon =\sum_{i=1}^{n}\varepsilon _{i}$, and
with ${\cal{A}}$ already defined, we obtain the following
non linear minimization problem:
\begin{multline}
\min_{q,t} (q^{T}({\cal{A}}+{\cal{B}})q + t^{T}{\cal{C}}t + \delta t + \varepsilon Q(q)^{T}W(q)t +  \\
		    	        \lambda (1-q^{T}q)^{2})
\label{eq:Randt-errorfunction}
\end{multline}
which is the sum of 5 terms. The number of parameters to be estimated
is 7 (4 for the unit quaternion and 3 for the translation).
It is worthwhile to notice that the
number of terms of this error function is constant with respect to $n$, i.e.,
the number of hand-eye motions. For such minimization problems one may use {\em constrained step methods}
such as the trust region method \citep{Fletcher90}, \citep{kn:Ya}.

\section{Stability analysis and method comparison}
One of the most important merits of any hand-eye calibration method is its
stability with respect to various perturbations. There are two main sources of
perturbations: errors associated with camera calibration and errors associated
with the robot motion. Indeed, the parameters of both the direct and
inverse kinematic models of robots are not perfect. As a consequence the real
motions associated with both the hand and the camera are known up to some
uncertainty. It follows that the estimation of the hand-eye transformation has
errors associated with it and it is important to quantify these errors in
order to determine the stability of a given method and to compare various
methods.

In order to perform this stability analysis we designed a 
stability analysis based on the following grounds:
\begin{itemize}
\item Nominal values for the parameters of 
the hand-eye transformation ($X$ or $Y$) are
provided;
\item Also are provided $n+1$ matrices $A_1$, \ldots $A_{n+1}$ from which $n$ 
hand motions can be computed either with (see Section 2):
\[	B_{i i+1} = X^{-1} A_{i+1} A_{i}^{-1} X		\]
or with:
\[	B_{1 i} = Y^{-1} A_{1}^{-1} A_i Y 	\]
\item Gaussian noise or uniform noise is added to both the camera and hand
motions and $X$ (or $Y$) is estimated in the presence of this noise using
three different methods: Tsai-Lenz, closed-form solution, and non-linear
optimization, and
\item We study the variations of the estimation of the hand-eye transformation
as a function of the noise being added and/or as a function of the number of
motions ($n$).
\end{itemize}

Since both rotations and translations may be represented as vectors, adding
noise to a transformation consists of adding random numbers to each one of the
vectors' components. Random numbers simulating noise are obtained using a
random number generator either with a uniform distribution in the interval 
$[-C/2, +C/2]$, or with a Gaussian distribution with a standard deviation
equal to $\sigma$. Therefore
the level of noise that is added is associated either with
the value of $C$ or with the value of $\sigma$ (or, more precisely, with the
value of 2$\sigma$). In what follows the level of noise is in fact represented
as a ratio: the values of the actual random numbers divided by the values 
of the perturbed parameters.

In the case of a rotation, the vector associated with this rotation has a
module equal to 1 and therefore the ratio is simply either $C$ or 2$\sigma$.
In the case of a translation the ratio is computed with respect to a nominal
value estimated over all the perturbed translations:
\[
\| t_{nominal} \| = 
\frac{ \sum_{i=1}^n ( \| t_{A_{i i+1}} \| + \| t_{B_{i i+1}} \| ) }{2n}
\]
where $t_{A_{i i+1}}$ is the translation vector associated with $A_{i i+1}$.

For each noise level and for a large number $J$ of trials we compute the
errors associated with rotation and translation as follows:
\[
e_{rot} = \sqrt{ \frac{1}{J} \sum _{j=1}^{J} \| \tilde{R}_{j} - R \| ^{2} }
\]
and:
\[
e_{tr} = \frac{ \sqrt{\frac{1}{J} \sum _{j=1}^{J} \| \tilde{t}_{j} - t \| ^{2}}}	      {\| t \|}
\]
where $R$ and $t$ are the nominal values of the transformation being estimated
($X$ or $Y$), $\tilde{R}_j$ and $\tilde{t}_j$ are the estimated rotation and
translation for some trial $j$, and $J$ is the number of trials for each noise
level (defined either by $C$ or by $\sigma$). In all our experiments we set
$J=1000$ and $\| t\| = 157$mm.
\begin{figure}
%\centerline{\psfig{figure=../fadi/pince/data/trace/rot.u.r.ps,width=9.5cm}}
\centering
\includegraphics[width=0.5\textwidth]{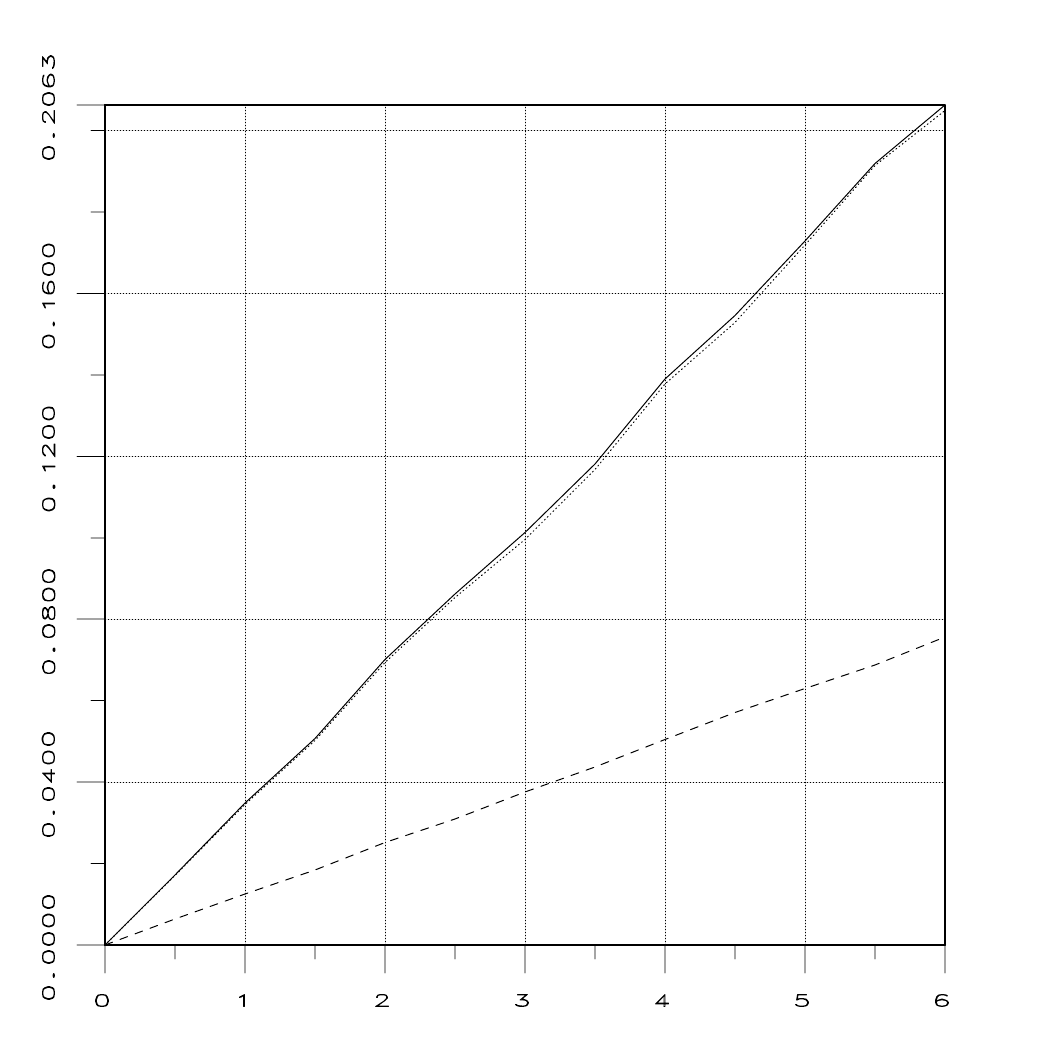}
\caption{Error in rotation in the presence of uniform noise perturbing the
rotation axes.}
\label{fig:erreurs1}
\end{figure}

\begin{figure}
%\centerline{\psfig{figure=../fadi/pince/data/trace/tra.u.r.ps,width=9.5cm}}
\centering
\includegraphics[width=0.5\textwidth]{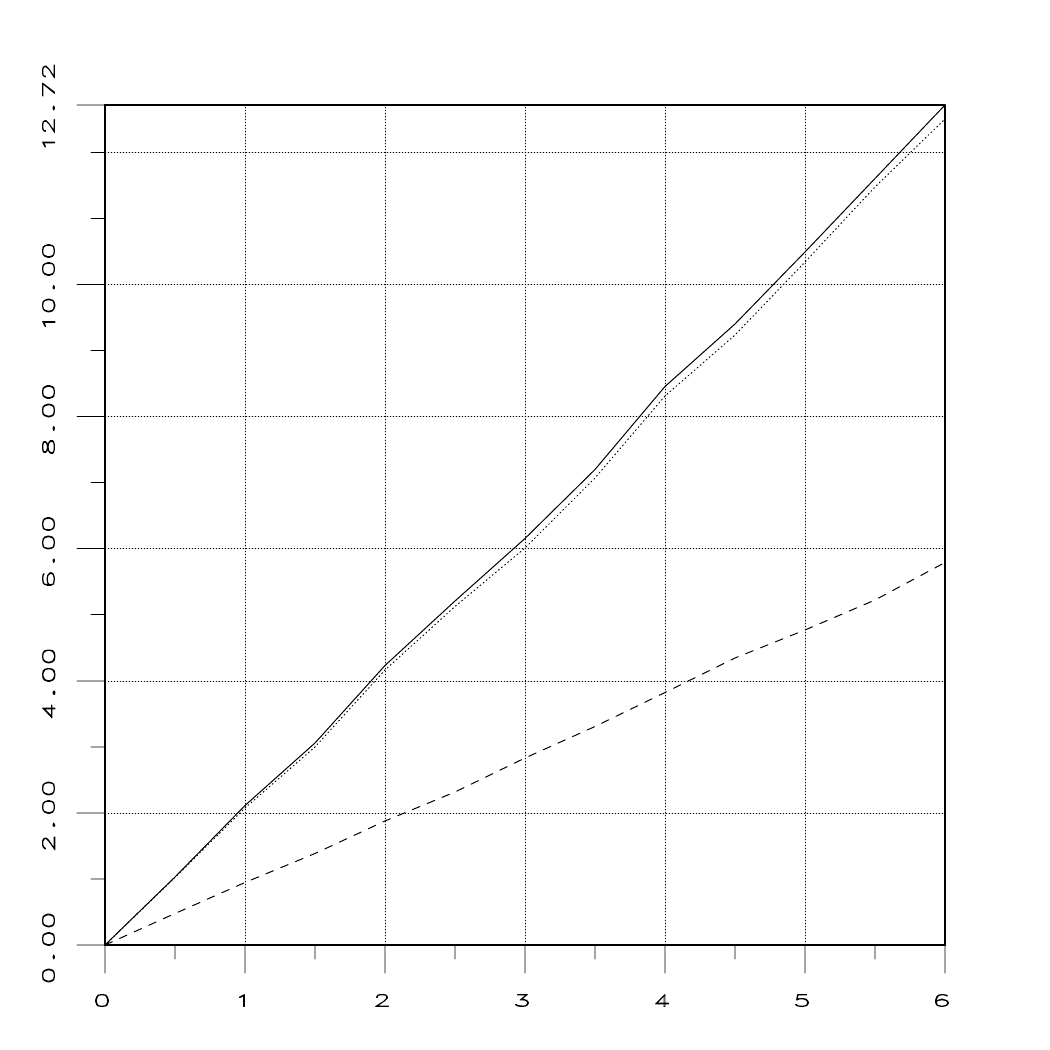}
\caption{Error in translation in the presence of uniform noise perturbing the
rotation axes.}
\label{fig:erreurs2}
\end{figure}

\begin{figure}
%\centerline{\psfig{figure=../fadi/pince/data/trace/rot.u.rt.ps,width=9.5cm}}
\centering
\includegraphics[width=0.5\textwidth]{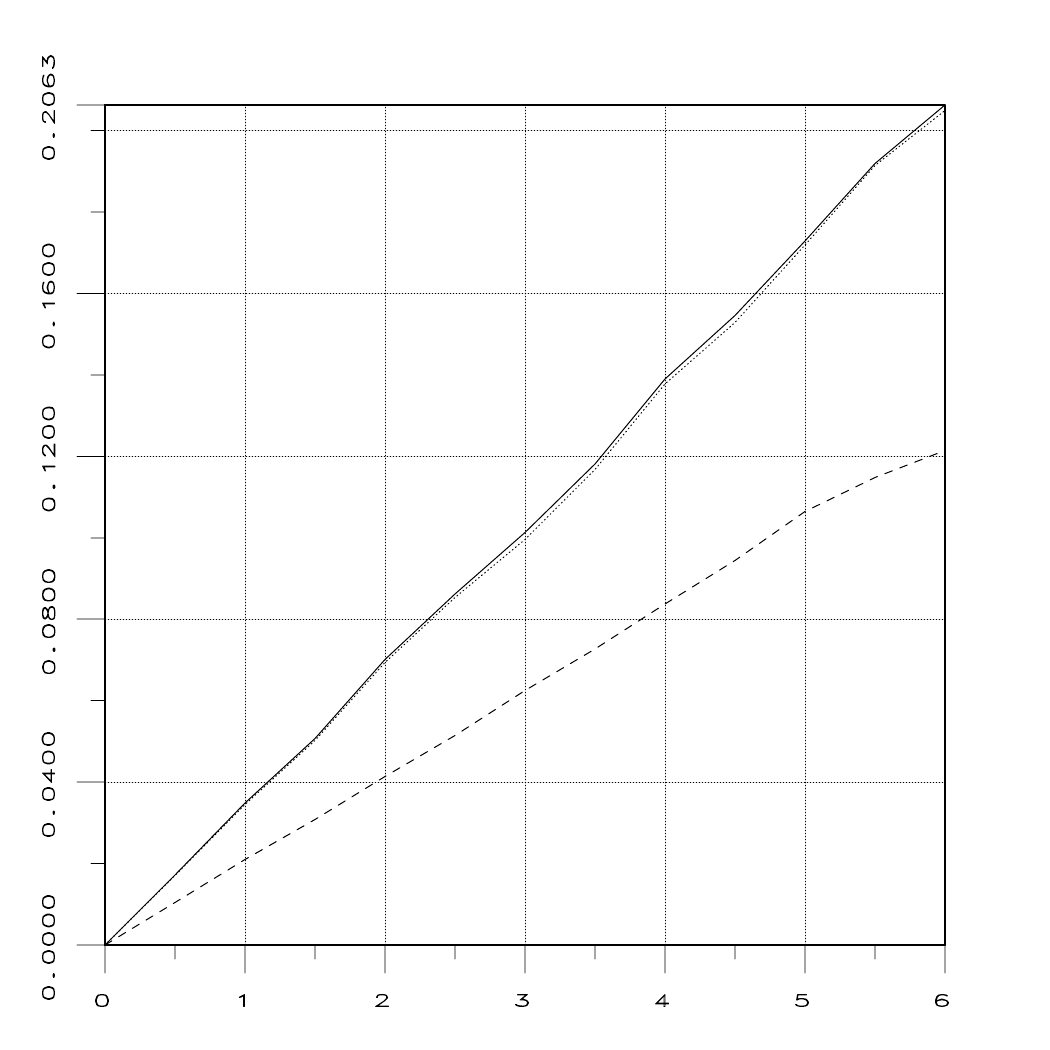}
\caption{Error in rotation in the presence of uniform noise perturbing the
translation vectors and the rotation axes.}
\label{fig:erreurs3}
\end{figure}

\begin{figure}
%\centerline{\psfig{figure=../fadi/pince/data/trace/tra.u.rt.ps,width=9.5cm}}
\centering
\includegraphics[width=0.5\textwidth]{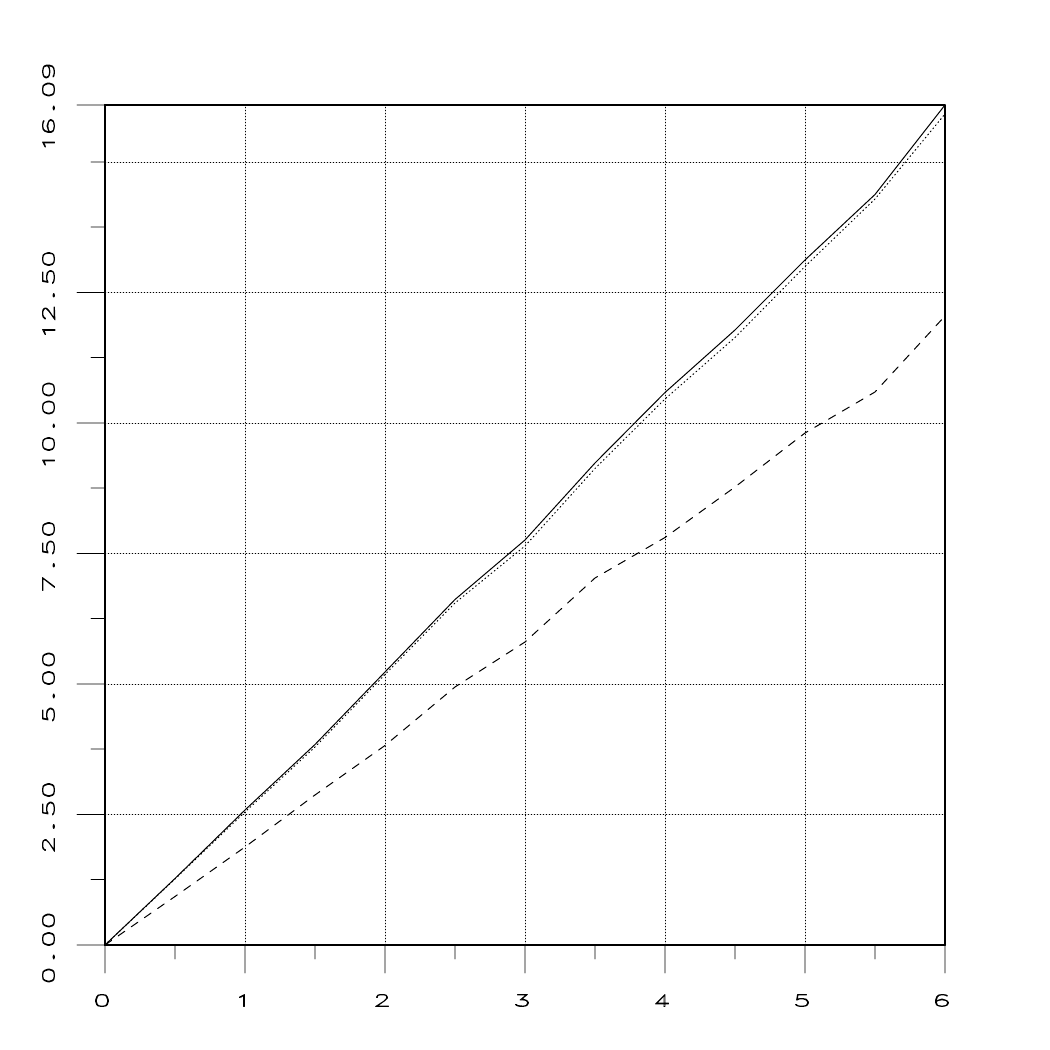}
\caption{Error in translation in the presence of uniform noise perturbing the
translation vectors and the rotation axes.}
\label{fig:erreurs4}
\end{figure}

\begin{figure}
%\centerline{\psfig{figure=../fadi/pince/data/trace/rot.g.r.ps,width=9.5cm}}
\centering
\includegraphics[width=0.5\textwidth]{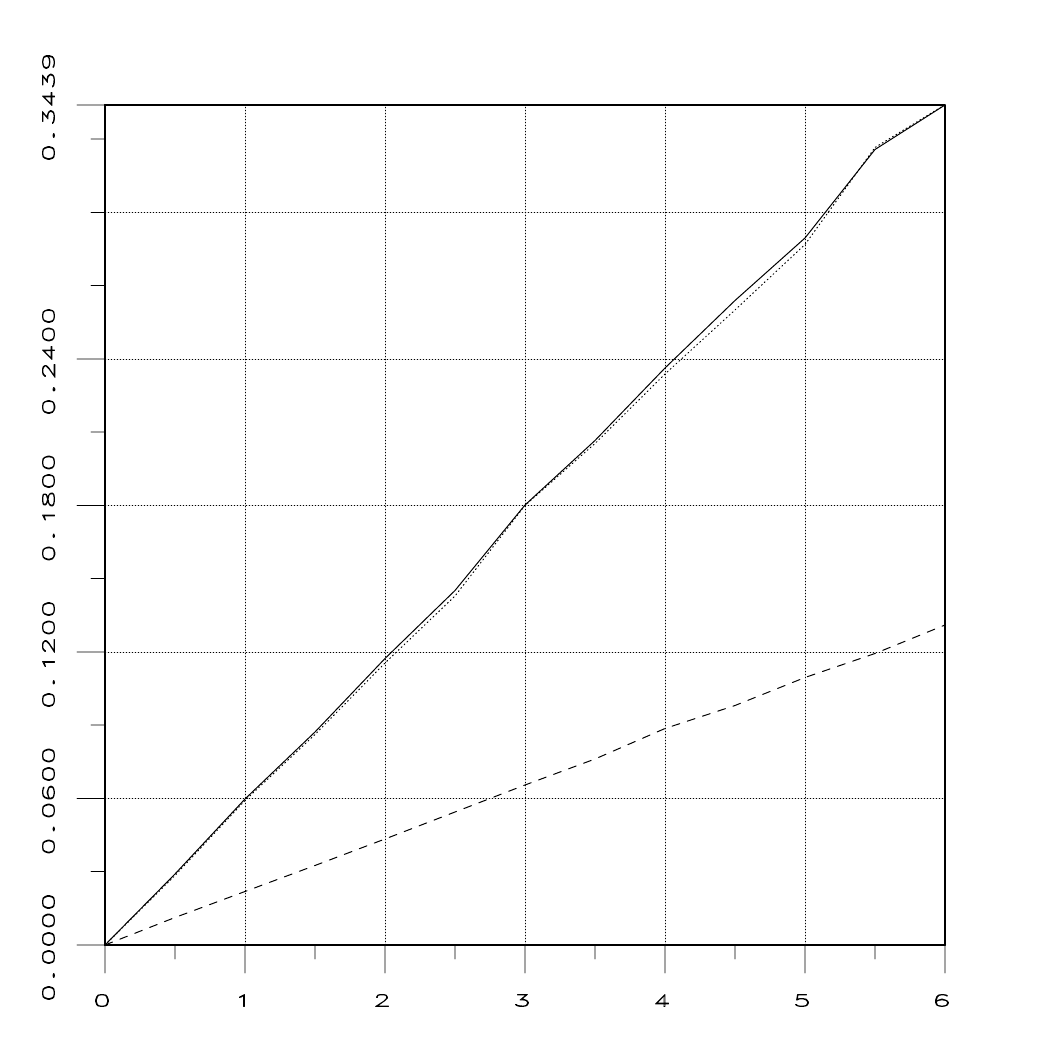}
\caption{Error in rotation in the presence of Gaussian noise perturbing the
rotation axes.}
\label{fig:erreurs5}
\end{figure}

\begin{figure}
%\centerline{\psfig{figure=../fadi/pince/data/trace/tra.g.r.ps,width=9.5cm}}
\centering
\includegraphics[width=0.5\textwidth]{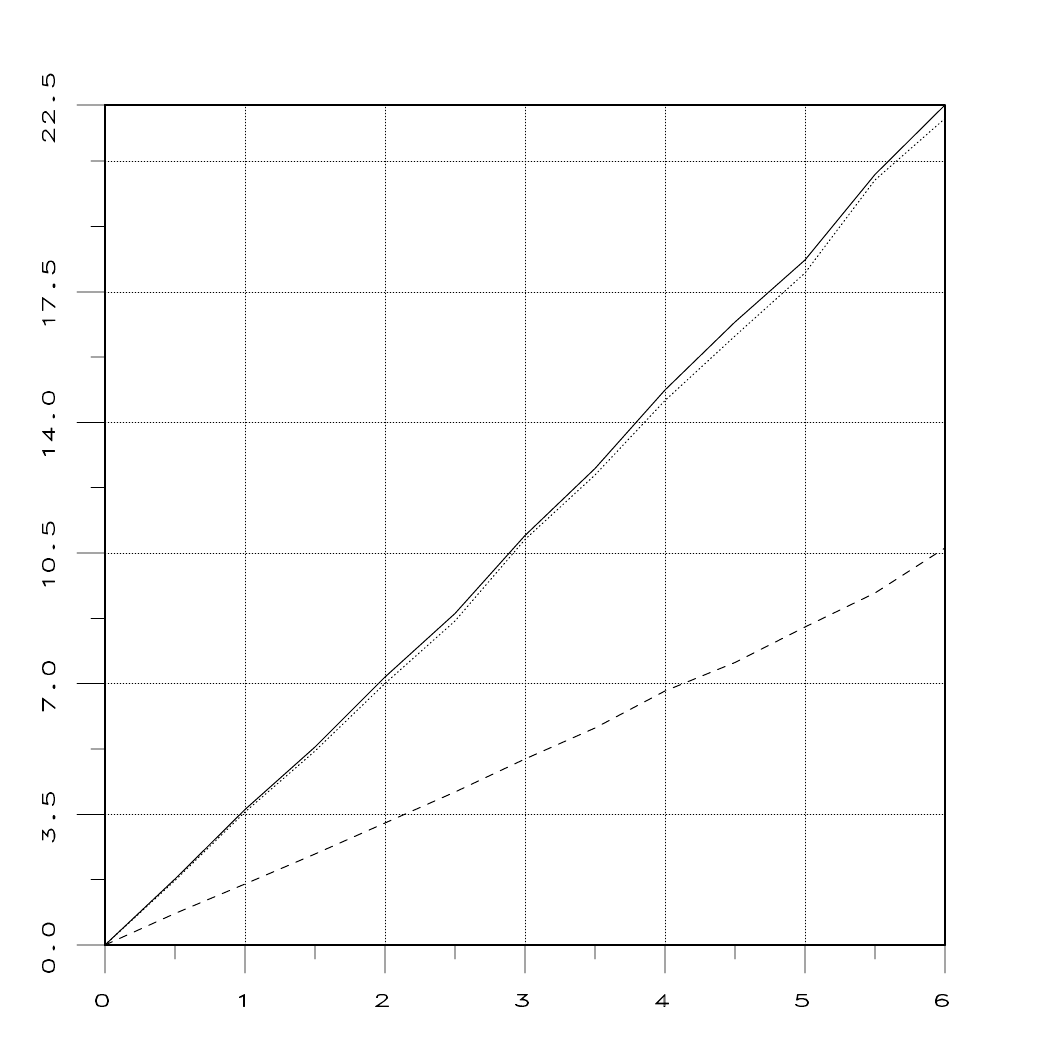}
\caption{Error in translation in the presence of Gaussian noise perturbing the
rotation axes.}
\label{fig:erreurs6}
\end{figure}

\begin{figure}
%\centerline{\psfig{figure=../fadi/pince/data/trace/rot.g.rt.ps,width=9.5cm}}
\centering
\includegraphics[width=0.5\textwidth]{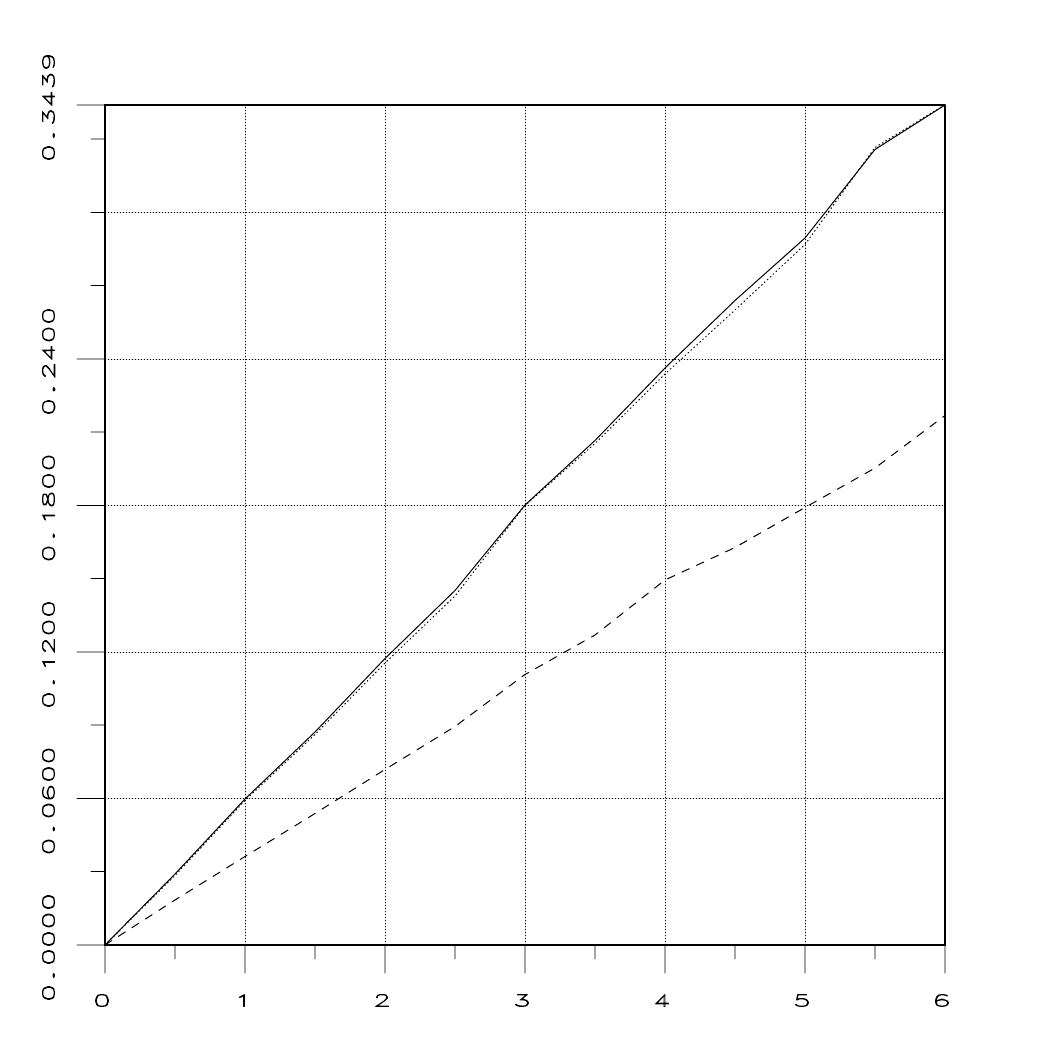}
\caption{Error in rotation in the presence of Gaussian noise perturbing the
translation vectors and the rotation axes.}
\label{fig:erreurs7}
\end{figure}

\begin{figure}
%\centerline{\psfig{figure=../fadi/pince/data/trace/tra.g.rt.ps,width=9.5cm}}
\centering
\includegraphics[width=0.5\textwidth]{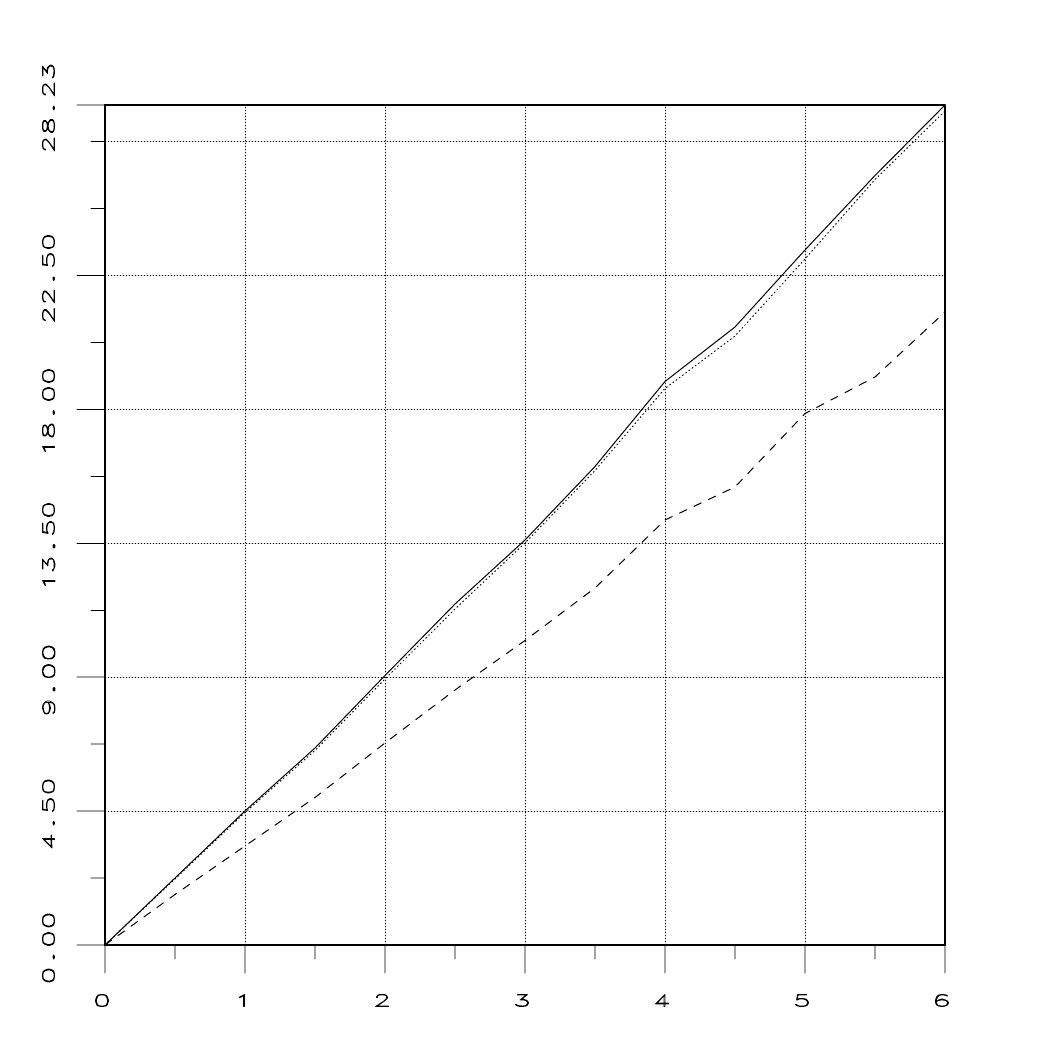}
\caption{Error in translation in the presence of Gaussian noise perturbing the
translation vectors and the rotation axes.}
\label{fig:erreurs8}
\end{figure}

\begin{figure}
%\centerline{\psfig{figure=../fadi/pince/data/trace/rot.rt.racine.ps,width=9.5cm}}
\centering
\includegraphics[width=0.5\textwidth]{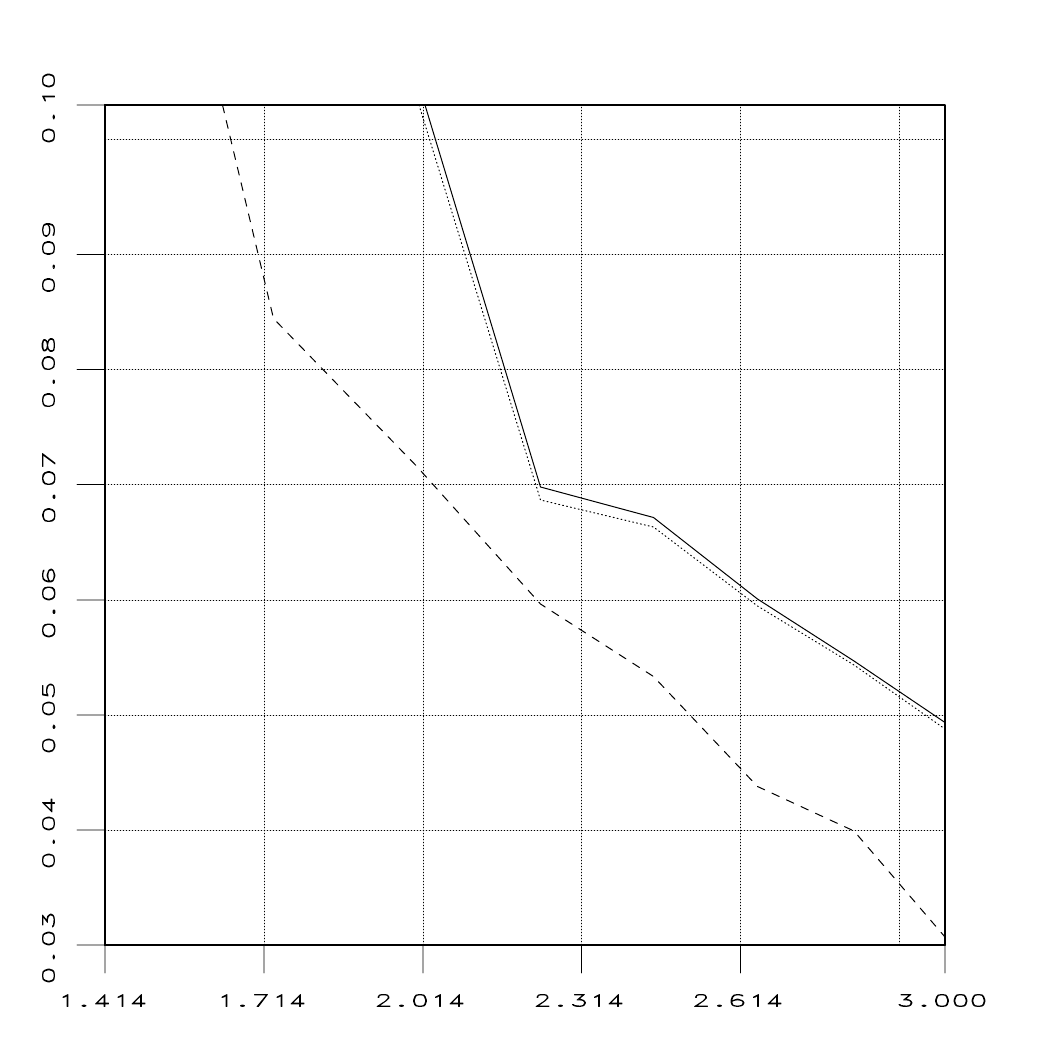}
\caption{Error in rotation as a function of the number of motions.}
\label{fig:erreurs9}
\end{figure}

\begin{figure}
%\centerline{\psfig{figure=../fadi/pince/data/trace/tra.rt.racine.ps,width=9.5cm}}
\centering
\includegraphics[width=0.5\textwidth]{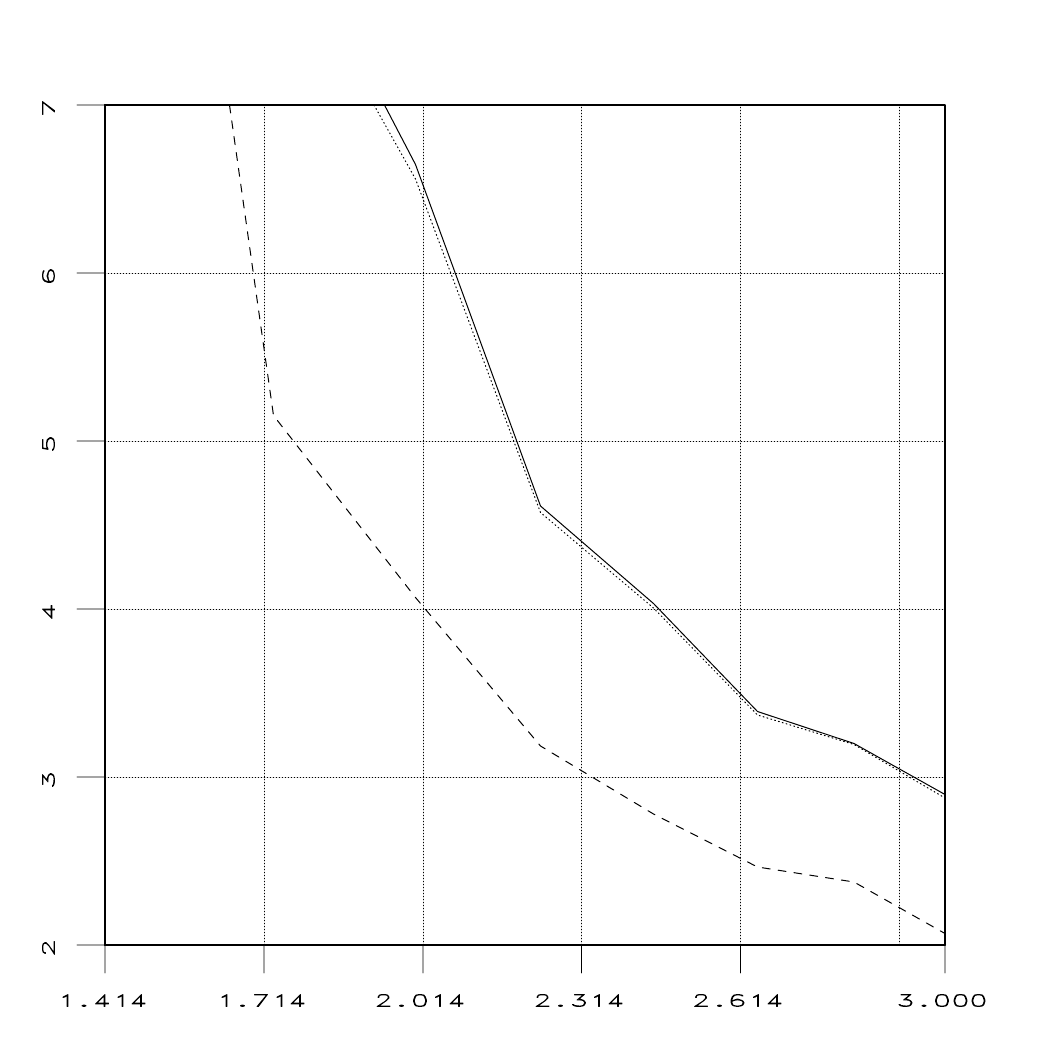}
\caption{Error in translation as a function of the number of motions.}
\label{fig:erreurs10}
\end{figure}

The following figures show the above errors as a function of the percentage of
noise. The percentage of noise varies from 1\% to 6\%. The full curves (---)
correspond to the method of Tsai \& Lenz, the dotted curves ( \ldots )
correspond to the closed-form method and the dashed curves (- - -)
correspond to the non-linear method. Figure~\ref{fig:erreurs1}
through  Figure~\ref{fig:erreurs8} correspond to two motions ($n=2$) of the
hand-eye device while on Figure~\ref{fig:erreurs9} and 
Figure~\ref{fig:erreurs10} the number of motions varies from 2 to 9.

Figure~\ref{fig:erreurs1} and Figure~\ref{fig:erreurs2} show the rotation and
translation errors as a function of uniform noise added to the rotational part
of the hand and camera motions. Figure~\ref{fig:erreurs3} and
Figure~\ref{fig:erreurs4} show the rotation and translation errors as a
function of uniform noise added both to the rotational and translational parts
of the camera and hand motions. Figure~\ref{fig:erreurs5} through
Figure~\ref{fig:erreurs8} are similar to Figure~\ref{fig:erreurs1} through
Figure~\ref{fig:erreurs4} but the uniform distribution of the noise has been
replaced by a Gaussian distribution.

It is interesting to notice that the Tsai-Lenz method and closed-form 
method have almost the same behaviour while the non-linear method provides
more accurate results in all the situations. The fact that the results
obtained with the first two methods are highly correlated may be due to the
fact that both these methods decouple the estimation of rotation from the
estimation of translation. This behaviour seems to be independent 
with respect to the noise type (uniform or Gaussian) and of whether only
rotation is perturbed or rotation and translation are perturbed
simultaneously. We conclude that the decoupling of rotation and translation
introduces a bias in the estimation of the hand-eye transformation.

As other authors have done in the past, it is interesting to analyse the
behaviour of hand-eye calibration with respect to the number of motions. In
order to perform this analysis we have to fix the percentage of noise.
Figure~\ref{fig:erreurs9} and Figure~\ref{fig:erreurs10} show the rotational
and translational errors as a function of the squared root of the number of
motions ($\sqrt{n}$ varies from $1.414$ to $3$). The noise ratio has been
fixed to the worst case for rotations, e.g., 6\% and to 2\% for translations.
Both rotational and translational noise distributions are Gaussian. For
example, for 4 motions the error in translation is of 4\% for the non-linear
method and of 6.5\% for the other two methods.

\section{Experimental results}

\begin{table*}[t!h!]
\begin{center}
\begin{tabular}{||c|c|c||}
\hline
$AX=XB$ & $\sum \| R_{A}R_{X}-R_{X}R_{B} \|^{2}$ &
$\frac{\sum \| (R_{A} - I) t_{X} - R_{X} t_{B} + t_{A} \|^{2}}
      {\sum \| R_{X} t_{B} - t_{A} \|^{2}}$  \\
\hline
Tsai-Lenz     & 0.0006   & 0.032  \\
\hline
Closed-form solution & 0.0005   & 0.029  \\
\hline
Non-linear optimization & 0.0003  & 0.019 \\
\hline
\end{tabular}
\end{center}
\caption{The classical formulation used with the first data set.}
\label{table-1}
\end{table*}

\begin{table*}[t!h!]
\begin{center}
\begin{tabular}{||c|c|c||}
\hline
$AX=XB$ & $\sum \| R_{A}R_{X}-R_{X}R_{B} \|^{2}$ &
$\frac{\sum \| (R_{A} - I) t_{X} - R_{X} t_{B} + t_{A} \|^{2}}
      {\sum \| R_{X} t_{B} - t_{A} \|^{2}}$  \\
\hline
Tsai-Lenz     & 0.0014   & 0.036  \\
\hline
Closed-form solution & 0.0014   & 0.023  \\
\hline
Non-linear optimization & 0.0017  & 0.015 \\
\hline
$MY=M'YB$ & $\sum \| NR_{Y}-R_{Y}R_{B} \|^{2}$ &
$\frac{ \sum \| (N - I) t_{Y} - R_{Y} t_{B} + t_{N} \|^{2}}
      { \sum \| R_{Y} t_{B} - t_{N} \|^{2}}$  \\
\hline
Tsai-Lenz     & 0.0031   & 0.0021  \\
\hline
Closed-form solution & 0.0015   & 0.001  \\
\hline
Non-linear optimization & 0.0013  & 0.0006 \\
\hline
\end{tabular}
\end{center}
\caption{The classical and the new formulations used with the second data set.}
\label{table-2}
\end{table*}

\begin{table*}[t!h!]
\begin{center}
\begin{tabular}{||c|c|c|c||}
\hline
$AX=XB$ & $\sum \| R_{A}R_{X}-R_{X}R_{B} \|^{2}$ &
$\frac{\sum \| (R_{A} - I) t_{X} - R_{X} t_{B} + t_{A} \|^{2}}
      {\sum \| R_{X} t_{B} - t_{A} \|^{2}}$ & CPU time \\
\hline
Tsai-Lenz     & 0.014   & 0.23 & 0.08 \\
\hline
Closed-form solution & 0.036   & 0.223  & 0.06 \\
\hline
Non-linear optimization & 0.258  & 0.058 & 0.21 \\
\hline
$MY=M'YB$ & $\sum \| NR_{Y}-R_{Y}R_{B} \|^{2}$ &
$\frac{ \sum \| (N - I) t_{Y} - R_{Y} t_{B} + t_{N} \|^{2}}
      { \sum \| R_{Y} t_{B} - t_{N} \|^{2}}$  & \\
\hline
Tsai-Lenz     & 0.038   & 0.039  & 0.06 \\
\hline
Closed-form solution & 0.035   & 0.037  & 0.08 \\
\hline
Non-linear optimization & 0.04  & 0.034 & 0.25 \\
\hline
\end{tabular}
\end{center}
\caption{The classical and the new formulations used with the third data set.
The last column indicates the CPU time in seconds on a Sparc-10 Sun computer.}
\label{table-3}
\end{table*}

In this section we report some experimental results obtained with three sets of
data. The first data set was provided by Fran\c{c}ois Chaumette from IRISA and
the second and third data sets were obtained at LIFIA. 
The first data set was obtained
with 17 different positions of the hand-eye device with respect to a
calibrating object. The second data set was obtained with 7 such positions.
The third data set was obtained with 6 positions.
For the first set only the extrinsic camera parameters were provided while
for the latter sets we had access to the full 3$\times$4 perspective matrices.
Therefore, the latter sets allowed us to test both the classical and the new
formulations. The only restrictions imposed onto the robot motions are due to
the fact that in eachone of its positions the camera mounted onto the robot
must see the calibration pattern.

In order to calibrate the camera we used the method proposed by Faugeras \&
Toscani \citep{FaugerasToscani86} and the following setup. The calibrating
pattern consists of a planar grid of size 200$\times$300mm that can move
along an axis perpendicular to its plane. The distance from this calibrating
grid to the camera varies during hand-eye calibration between 600mm and 1000mm.
This calibration setup combined with the Faugeras-Toscani method provides very
accurate camera calibration data. This is mainly due to the accuracy of the
grid points (0.1mm), to the accuracy of point localization in the image (0.1
pixels), and to the large number of calibrating points being used (460
points). Moreover, camera calibration errors can be
neglected with respect to robot calibration errors (see below).

Since the two formulations are mathematically equivalent, we have been able to
test and compare the classical Tsai-Lenz method with the two methods
developed in this paper. Table~\ref{table-1}, Table~\ref{table-2}, and
Table~\ref{table-3} summarize the results obtained with the three data sets
mentioned above. The lengths of the translation vectors thus obtained are:
$\| t_{X}\| = 93$mm and $\| t_{Y}\| = 681$mm.

The second columns of these tables show the sum of squares of the absolute error in rotation.
The third columns show the sum of squares of the relative error in translation.

These experimental results seem to confirm that, on one hand, the non-linear
method provides a better estimation  of the translation vector -- {\em at the
cost of a, sometimes, slightly less accurate rotation} -- and, on the other
hand, the new formulation provides a better estimation of the transformation
parameters than the classical formulation.

It is worthwhile to notice that, while the non-linear technique provides the
most accurate results with simulated data, the linear and
closed-form techniques provide sometimes a better estimation of rotation with
real data. This is due to the fact that the robot's kinematic chain is not
perfectly calibrated and therefore there are errors associated with the
robot's translation parameters. Obviously, these errors do not obey the noise
models used for simulations. The linear and closed-form techniques estimate
the rotation parameters independently of the robot's translation
parameters and therefore the rotation thus estimated is not affected by
translation errors. However, in practice we prefer the non-linear technique.

\section{Discussion}
In this paper we attacked the problem of hand-eye calibration. In addition to
the classical formulation, i.e., $AX=XB$, we suggest a new formulation that
directly uses the 3$\times$4 perspective matrices available with camera
calibration: $MY=M'YB$. The advantage of the new formulation with
respect to the classical one is that it avoids the decomposition of the
perspective matrix into intrinsic and extrinsic camera parameters. Indeed, it
has long been recognised in computer vision research that this decomposition
is unstable.

Moreover, we show that the new formulation has a mathematical structure that
is identical with the mathematical structure of the classical formulation. The
advantage of this mathematical analogy is that, the two formulations being 
variations of the same one, any method for solving the problem applies
to both formulations.

We develop two resolution methods, the first one solves for rotation and then for translation
while the second one solves simultaneously for rotation and translation. Using
unit quaternions to represent rotations, the first method leads to a closed
form solution introduced by Faugeras \& Hebert \citep{FaugerasHebert86}
while the second one is new and leads to non-linear optimization. Among
the many robust non-linear optimization methods that are available, we chose
the Levenberg-Marquardt technique.

Both the stability analysis and the results obtained with experimental data
from two laboratories show that the non-linear optimization method yields the
most accurate results. Linear algebra techniques (the Tsai-Lenz method) and
the closed-form method (using unit quaternions) are of comparable accuracy. 

The new formulation provides
much more accurate hand-eye calibration results than the classical formulation. This
improvement in accuracy seems to confirm that the decomposition of the perspective 
matrix into intrinsic and extrinsic parameters introduces some errors.
Nevertheless, the intrinsic parameters, even if they don't need to be made
explicit, are assumed to be constant during calibration. We are perfectly
aware that this assumption is not very realistic and may cause problems in
practice. We are currently investigating ways to give up this assumption.

Also, we investigate ways to perform hand-eye calibration and robot
calibration simultaneously. Indeed, in many applications such as nuclear and
space environments it may be useful to calibrate a robot simply by calibrating
a camera mounted onto the robot.

\appendix
\section{Rotation and unit quaternion}
The use of unit quaternions to represent rotations is justified by an elegant
closed-form solution associated with the problem of optimally estimating
rotation from 3-D to 3-D vector correspondences \citep{FaugerasHebert86},
\citep{Horn87-quat}, \citep{SabataAggarwal92}, \citep{HoraudMonga93}. 
In section \ref{section:unified-optimal-solution} we 
stressed the similarity between the hand-eye calibration problem and the problem
of optimally estimating the rotation between sets of 3-D features. In this
appendix we briefly recall the definition of quaternions, some useful
properties of the quaternion multiplication operator, and the relationship
between 3$\times$3 orthogonal matrices and unit quaternions. 

A quaternion is a 4-vector that may be viewed as a special case of complex
numbers that have one real part and three imaginary parts:
\[
q = q_{0} + i q_{x} + j q_{y} + k q_{z}
\]
with:
\[ i ^{2} = j ^{2} = k^{2} = ijk = -1 \]

One may define quaternion multiplication (denoted by $\ast$) as follows:
\[
r \ast q = (r_{0} + i r_{x} + j r_{y} + k r_{z})
                       (q_{0} + i q_{x} + j q_{y} + k q_{z})
\]
which can be written using a matrix notation:
\[
r \ast q = Q(r) q = W(q) r
\]
with:
\[
Q(r) = \left( \begin{array}{rrrr}
r_{0} & -r_{x} & -r_{y} & -r_{z} \\
r_{x} &  r_{0} & -r_{z} &  r_{y} \\
r_{y} &  r_{z} &  r_{0} & -r_{x} \\
r_{z} & -r_{y} &  r_{x} &  r_{0}
           \end{array}
    \right)
\]
and:
\[
W(r) = \left( \begin{array}{rrrr}
r_{0} & -r_{x} & -r_{y} & -r_{z} \\
r_{x} &  r_{0} &  r_{z} & -r_{y} \\
r_{y} & -r_{z} &  r_{0} &  r_{x} \\
r_{z} &  r_{y} & -r_{x} &  r_{0}
           \end{array}
    \right)
\]
One may easily verify the following properties:
\begin{align*}
Q(r)^{T} Q(r) &=  Q(r)Q(r)^{T} = r^{T}rI
\\
W(r)^{T} W(r) &=  W(r)W(r)^{T} = r^{T}rI
\\
Q(r)q &= W(q) r \\
Q(r)^{T} r &= W(r)^{T}  r = r^{T}r e\\
Q(r) Q(q) &= Q(Q(r)q) \\
W(r) W(q) &= W(W(r)q) \\
Q(r) W(q)^{T} &= W(q)^{T} Q(r)
\end{align*}
$e$ being the unity quaternion: $e=(1 \; 0 \; 0 \; 0)$.

The dot-product of two quaternions is:
\[
r \cdot q = r_{0}q_{0} + r_{x}q_{x} +  r_{y} q_{y} + r_{z} q_{z}
\]
The conjugate quaternion of $q$, $\overline{q}$ is defined by:
\[
\overline{q} = q_{0} - i q_{x} - j q_{y} - k q_{z}
\]
and obviously we have:
\[
q \ast \overline{q} = q \cdot q = \| q \|^{2}
\]
An interesting property that is straightforward and which will be used in the
next section is:
\begin{equation}
\| r \ast q \|^{2}  =  \|  r \|^{2} \|  q \|^{2}
\label{eq:r*q=r2q2}
\end{equation}

A 3-vector may well be viewed as a purely imaginary quaternion (its real part
is equal to zero). One may notice that $W(v)$ and $Q(v)$ associated with a
3-vector $v$ are skew-symmetric matrices.

Let $q$ be a unit quaternion, that is $q\cdot q=1$, and let $v$ be a purely
imaginary quaternion. We have:
\begin{eqnarray}
\label{eq:rotation-quaternion}
v' & = & q \ast v \ast \overline{q} \nonumber \\
   & = & (Q(q) v) \ast \overline{q} \\
   & = & ( W(q)^{T} Q(q)) v \nonumber
\end{eqnarray}
and one may easily figure out that:
\begin{multline*}
W(q)^{T}Q(q) = \left( \begin{matrix}
1 & 0 \\
0 & q_{0}^{2}+q_{x}^{2}-q_{y}^{2}-q_{z}^{2} \\
0 & 2(q_{x}q_{y}+q_{0}q_{z}) \\ 
0 & 2(q_{x}q_{z}-q_{0}q_{y}) 
\end{matrix} \right. \\
\left. \begin{matrix}
0 & 0 \\
2(q_{x}q_{y}-q_{0}q_{z}) & 2(q_{x}q_{z}+q_{0}q_{y}) \\
q_{0}^{2}-q_{x}^{2}+q_{y}^{2}-q_{z}^{2} & 2(q_{y}q_{z}-q_{0}q_{x}) \\
2(q_{y}q_{z}+q_{0}q_{x}) & q_{0}^{2}-q_{x}^{2}-q_{y}^{2}+q_{z}^{2}
\end{matrix} \right)
\end{multline*}
is an orthogonal matrix. Hence, $v'$ given by
eq.~(\ref{eq:rotation-quaternion}) is a 3-vector (a purely imaginary
quaternion) and is the image of $v$ by a rotation transformation $R$:
\begin{multline*}
R =
\left( \begin{matrix}
q_{0}^{2}+q_{x}^{2}-q_{y}^{2}-q_{z}^{2} & 2(q_{x}q_{y}-q_{0}q_{z}) \\
2(q_{x}q_{y}+q_{0}q_{z}) & q_{0}^{2}-q_{x}^{2}+q_{y}^{2}-q_{z}^{2} \\
2(q_{x}q_{z}-q_{0}q_{y}) & 2(q_{y}q_{z}+q_{0}q_{x}) 
\end{matrix} \right. \\
\left. \begin{matrix}
2(q_{x}q_{z}+q_{0}q_{y}) \\
2(q_{y}q_{z}-q_{0}q_{x}) \\
q_{0}^{2}-q_{x}^{2}-q_{y}^{2}+q_{z}^{2}
\end{matrix} \right)
\end{multline*}

\section{Derivation of equation \eqref{eq:f2-quat-transl}}
%(\ref{eq:f2-quat-transl})}

The expression of $f_{2}(q,t)$, i.e., equation~(\ref{eq:f2-quat-transl}), can be easily derived using
the properties of $W(q)$ and $Q(q)$ outlined in section~5:
\begin{eqnarray*}
\lefteqn{\| q \ast p_{i} - (K_{i}-I) t \ast q - p'_{i} \ast q \|^{2} } \\
 & & = \left( W(p_{i}) q - W(q)(K_{i}-I)t - Q(p'_{i})q \right) ^{T} \\
  & &   \left( W(p_{i}) q - W(q)(K_{i}-I)t - Q(p'_{i})q \right) \\
 & & = q^{T} {\cal{B}}_{i} q + t^{T} {\cal{C}}_{i} t + {\cal{D}}_{i} t - 2 q^{T} W(p_{i})^{T}W(q)(K_{i}-I)t
\end{eqnarray*}
where the expressions of ${\cal{B}}_{i}$, ${\cal{C}}_{i}$, and $\delta _{i}$ are:
\begin{eqnarray*}
{\cal{B}}_{i} & = & (p_{i}^{T}p_{i} + p_{i}'^{T}p_{i}')I - W(p_{i})^{T}Q(p_{i}') -
            Q(p_{i}')^{T}W(p_{i}) \\
{\cal{C}}_{i} & = & K_{i}^{T}K_{i} - K_{i} - K_{i}^{T} + I \\
\delta _{i} & = & 2p_{i}'^{T}(K_{i} - I)
% \cal{E}_{i} & = & -2p_{i}^{T}(R_{B_{i}} - I)
\end{eqnarray*}
The last term may be transformed as follows:
\begin{multline*}
 - 2 q^{T} W(p_{i})^{T}W(q)(K_{i}-I)t \\
  =  - 2 p_{i}^{T}Q(q)^{T}W(Q)(K_{i}-I)t \\
  =  - 2p_{i}^{T}\left( W(q)^{T}Q(q) \right) ^{T} (K_{i}-I)t
\end{multline*}

The matrix $W(q)^{T}Q(q)$ is the unknown rotation 
and is equal to either $R_{X}$ or $R_{Y}$. The matrix $K_{i}$ is a rotation as well and
is equal to 
either $R_{A_{i}}$ or $N_{i}$. Notice that we have from equations~(\ref{eq:AX=XB:rotation}) and
(\ref{eq:NRY=RYRB}):
\begin{align*}
R_{X}^{T}R_{A_{i}} & =  R_{B_{i}} R_{X}^{T} \\
R_{Y}^{T}N_{i} & =  R_{B_{i}} R_{Y}^{T}
\end{align*}
Therefore one may write:
\[
\left( W(q)^{T}Q(q) \right) ^{T} (K_{i}-I) = \left( R_{B_{i}} - I \right) \left( W(q)^{T}Q(q) \right)
^{T}
\]
Finally we obtain for the last term:
\begin{align*}
-  2 q^{T} W(p_{i})^{T} & W(q)(K_{i}-I)t \\
 = & - 2p_{i}^{T}\left( W(q)^{T}Q(q) \right) ^{T} (K_{i}-I)t \\
 = &  - 2p_{i}^{T} (R_{B_{i}} - I) \left( W(q)^{T}Q(q) \right) ^{T} t \\
 = &  - 2p_{i}^{T} (R_{B_{i}} - I) Q(q) ^{T}   W(q) t
\end{align*}
and:
\[ \varepsilon _{i} = -2p_{i}^{T}(R_{B_{i}} - I) \]

\begin{acks}
 The authors acknowledge Roger Mohr, Long Quan,
Thai Quynh Phong,
Sth\'ephane Lavall\'ee, Fran\c{c}ois Chaumette, Claude Inglebert, Christian Bard,
Thomas Skordas, and Boguslaw Lorecki
for fruitful discussions and for their many insightful comments. Many thoughts
go to Patrick Fulconis and Christian Bard for the many hours they have spent
with the robot. Fran\c{c}ois Chaumette kindly provided robot and camera calibration data.
\end{acks}
\balance
\bibliographystyle{spbasic}
%\bibliography{general,stereo,horaud,tactile,books,phong}

\end{document}